\documentclass{article}

\usepackage{arxiv}
\usepackage{graphicx} 
\graphicspath{ {images/} }
\usepackage{subcaption}
\usepackage{amsmath}
\usepackage{url}
\usepackage[symbol, multiple]{footmisc}

\usepackage{verbatim}

\usepackage{siunitx}

\usepackage{ulem}

\renewcommand{\headeright}{}
\renewcommand{\undertitle}{}
\renewcommand{\shorttitle}{Time-coded spiking FT}

\usepackage{tikz}
\usepackage{textcomp}
\newcommand\copyrighttext{%
  \footnotesize \textcopyright 2022 IEEE. Personal use of this material is permitted.
  Permission from IEEE must be obtained for all other uses, in any current or future 
  media, including reprinting/republishing this material for advertising or promotional 
  purposes, creating new collective works, for resale or redistribution to servers or 
  lists, or reuse of any copyrighted component of this work in other works. \newline
  DOI: 10.1109/TC.2022.3162708, available at https://ieeexplore.ieee.org/document/9743682
  }
\newcommand\copyrightnotice{%
\begin{tikzpicture}[remember picture,overlay]
\node[anchor=south,yshift=10pt] at (current page.south) {\fbox{\parbox{\dimexpr\textwidth-\fboxsep-\fboxrule\relax}{\copyrighttext}}};
\end{tikzpicture}%
}

\begin{document}
\copyrightnotice

\title{Time-Coded Spiking Fourier Transform in Neuromorphic Hardware}

\date{}
\author{Javier~López-Randulfe\thanks{Correspondence author. E-mail: lopez.randulfe@tum.de} \thanks{Department of Informatics, Technical University of Munich, Munich, Germany} ,
        Nico Reeb\footnotemark[2] ,
        Negin Karimi\footnotemark[2] ,
        Chen Liu\thanks{Faculty of Electrical and Computer Engineering, Technische Universit{\"a}t Dresden, Dresden, Germany} ,
        Hector A. Gonzalez\footnotemark[3] ,
        Robin Dietrich\footnotemark[2] , \\
        \textbf{Bernhard Vogginger\footnotemark[3] ,}
        \textbf{Christian Mayr\footnotemark[3] ,}
        \textbf{and~Alois~Knoll\footnotemark[2]}
}


\maketitle

\begin{abstract}

After several decades of continuously optimizing computing systems, the Moore's law is reaching its end.
However, there is an increasing demand for fast and efficient processing systems that can handle large streams of data while decreasing system footprints.
Neuromorphic computing answers this need by creating decentralized architectures that communicate with binary events over time.
Despite its rapid growth in the last few years, novel algorithms are needed that can leverage the potential of this emerging computing paradigm
and can stimulate the design of advanced neuromorphic chips.
In this work, we propose a time-based spiking neural network that is mathematically equivalent to the Fourier transform.
We implemented the network in the neuromorphic chip Loihi and conducted experiments on five different real scenarios with an automotive frequency modulated continuous wave radar.
Experimental results validate the algorithm, and we hope they prompt the design of ad hoc neuromorphic chips that can improve the efficiency of state-of-the-art digital signal processors
and encourage research on neuromorphic computing for signal processing.

\end{abstract}


\section{Introduction}
\label{sec:introduction}
Spiking neural networks (SNNs) can be executed rapidly and with high energy efficiencies on dedicated neuromorphic hardware
owing to their inherent event-based operation and sparse communication.
Since the term was first coined \cite{mead1990neuromorphic}, there has been increasing interest in designing not only accurate neuron models describing the neurobiological dynamics in detail
but also hardware that natively supports event-based communication, sparse coding, and highly parallel brain-inspired operations with distributed memory. State-of-the-art neuromorphic chips \cite{furber2016large,thakur2018large}, such as SpiNNaker2\cite{YanLoihiSpinnaker22021} and Intel's Loihi \cite{davies2018loihi}, have shown remarkable performance in various tasks,
including event-based data processing, adaptive control, constrained optimization, and graph search \cite{davies2021advancing}. 

This offers a promising alternative to today's artificial intelligence (AI) systems, which build on the highly parallelized von Neumann computers (GPUs). Although these systems have become state-of-the-art solutions to many problems, including image classification and speech processing, they consume considerable energy. This introduces a limitation, e.g., for highly automated vehicles, where systems that process sensor data can drain more than $10 \%$ of the power stored for driving \cite{lin2018architectural}. The success of current AI algorithms further builds on the constant improvement of CPU/GPU capabilities. The improvement in the traditional von Neumann computers is, however, slowing down as we approach physical manufacturing limits. The imminent end of the Moore's law \cite{waldrop2016chips} indicates the necessity of exploring new computational technologies such as neuromorphic computing to further improve the performance and energy efficiency of next-generation AI algorithms.

Along with the continuous improvement of neuromorphic chips, SNN-based solutions have emerged in recent years for various applications and sensors,
ranging from speech recognition with resonate-and-fire neurons \cite{auge_end--end_2021}, object tracking for monocular vision \cite{piekniewski_unsupervised_2016, luo_siamsnn_2020},
object detection using raw temporal pulses of lidar sensors \cite{wang_temporal_2021} for lane keeping \cite{bing_end_2018},
feature extraction and motion perception \cite{paredes-valles_unsupervised_2020}, and collision avoidance  based on data obtained from a dynamic vision sensor \cite{salvatore_neuro-inspired_2020}. 
Currently, the most prominent task addressed in radar data processing using SNNs is gesture recognition \cite{kreutz_applied_2021, safa_improving_2021-1, tsang_radar-based_2021-1, yin_accurate_2021}.
Micro-Doppler signatures of hand movement are particularly suited for gestures and contain temporal information, which on the other hand leverages the recurrence ability of SNNs.
Recently, IMEC presented the \textmu Brain chip \cite{stuijt2021ubrain}, an event-driven, fully synthesizable architecture for SNNs, targeting low-power edge neuromorphic chips and applications such as radar-based hand-gesture recognition and image classification with MNIST.
This first successful demonstration of radar-based hand-gesture recognition substantiates the hypothesis of SNN superiority in terms of energy efficiency and computational execution time when applied to a suitable task using specifically designed neuromorphic hardware.


Among the wide variety of relevant signal processing algorithms that could benefit from neuromorphic efficiency, the Fourier transform (FT) represents an attractive choice.
The FT is not only the workhorse of modern signal processing but also governs almost every data-processing application in our digital age. Exploring a spike-based FT brings us closer to how biological organisms decode frequency tones
\cite{Carr3227} and follows the same motivation as the quantum research community working on quantum-based fast FT (FFT) implementations \cite{coppersmith2002approximate}, which is extending the application of core inference engines in specialized hardware (e.g., quantum or neuromorphic) to extracting frequency-based features.

In a specific case of radar processing, an efficient and accurate spiking alternative for the FT opens the door to implementing a full neuromorphic processing pipeline,
which would lead to using only one chip for handling the different tasks at hand.

This study proposes an alternative and novel spike-based FT (S-FT), which is suitable for neuromorphic hardware. This study is a major extension of the work presented in  \cite{lopez2021spiking}. The main novel aspects are listed below:

\begin{itemize}
    \item We introduce a novel time-based neuron model,
    that requires only one spike per input for computing a matrix multiplication.
    \item We introduce a novel sparse spiking network architecture that can replicate
    the structure of the FFT.
    \item We implement the spiking algorithm on the Loihi neuromorphic hardware \cite{davies2018loihi}
    and benchmark the algorithm on real-life scenarios from an automotive radar.
\end{itemize}

The results indicate that the proposed algorithm has competitive accuracy,
showing a low error throughout the entire spectrum of the processed data.
Although the estimated energy consumption is higher than state-of-the-art FFT accelerators,
we believe that next generations of neuromorphic hardware will
close this gap, making the S-FT a viable replacement for traditional versions of this algorithm.

The remainder of this study is organized as follows. Section II details the customized spiking neural model, the network architecture, and the target neuromorphic implementation.
Section III explains the validation experiments and the comparison framework.
Section IV discusses the implications of the obtained results,
and Section V concludes the paper.

\section{Spiking neural network}

We propose a spiking neuron model that can replicate matrix multiplications using time coding,
i.e., by representing each value with a single spike in time.
%
%
Namely, we use time-to-first-spike encoding to represent real numbers.
For a real value $x_i$ ranging from $x_\text{min}$ to $x_\text{max}$, the equivalent spike time will be $t_j$ for a time
range between $t_\text{min}$ and $t_\text{max}$:

\begin{equation}
    \label{eq:ttfs}
    t_j = t_\text{min} + \dfrac{t_\text{max}-t_\text{min}}{x_\text{max}-x_\text{min}} \cdot (x_\text{max}- x_j) \;.
\end{equation}

Equation (\ref{eq:ttfs}) can be simplified for the common case where $t_\text{min} =0$ and $x_\text{min} = - x_\text{max}$:
\begin{equation}
    \label{eq:ttfs_simplified}
    t_j = \dfrac{t_\text{max}}{2x_\text{max}} \cdot (x_{max}- x_j)  \;.
\end{equation}
This assumption is applied in the rest of the paper.
In the rest of the section, we will represent (\ref{eq:ttfs_simplified}) as
\begin{equation}
    \label{eq:ttfs_simplified_gamma}
    t_j = \gamma \cdot (x_\text{max} - x_j) \,,
\end{equation}
where $\gamma$ is the constant factor
\begin{equation}
    \label{eq:gamma}
    \gamma = \dfrac{t_\text{max}}{2 x_\text{max}} \,.
\end{equation}

\subsection{Neuron model}

The voltage dynamics of our model are the same as in the model proposed in \cite{rueckauer2018conversion}.
At a certain time $t$, the voltage of neuron $i$ depends only on the weights of the input neurons that
have already spiked, also called causal neurons and represented by $\Gamma_i^{<}$.
The contribution of each causal neuron $j \in \Gamma_i^{<}$ to the voltage of neuron $i$ is directly proportional
to the time elapsed since $j$ spiked. Thus, the voltage of neuron $i$ takes the form of
\begin{equation}
    \label{eq:ttfs_membrane_update}
    u_i(t) = \sum_{j \in \Gamma_i^{<}} w_{ij} (t-t_j) \;,
\end{equation}
where $w_{ij}$ is the weight of the synapse connecting $i$ with $j$, and each input neuron $j$ is restricted to produce one single spike.
Moreover, the voltage change between two consecutive steps results in the equation

\begin{equation}
    \label{eq:ttfs_membrane_variation}
    \Delta u_i = \Delta t \sum_{j \in \Gamma_i^{<}} w_{ij} \;, 
\end{equation}
and the neuron generates a spike whenever the voltage reaches a threshold voltage $u_{th}$.
We modify (\ref{eq:ttfs_membrane_update}) to obtain a model that can represent a linear combination $z = \textbf{W} \cdot x$,
where the element $i$ of the resulting vector $z$ takes the form
\begin{equation}
    \label{eq:linear_combination}
    z_i = \sum_j^N w_{ij} x_j \,.
\end{equation}
%


The discrete FT (DFT) is a specific implementation of (\ref{eq:linear_combination}), as it can be represented by the matrix-vector multiplication

\begin{equation}
    \label{eq:dft}
    y_k = \sum_{n=0}^{N-1} x_n \cdot \left[ cos\left(\dfrac{2 \pi}{N} kn\right) - i \cdot sin\left(\dfrac{2 \pi}{N} kn\right)   \right] \,,
\end{equation}
%
%
where $y_k$ represents the result of the \textit{k-th} bin of the FT over the input signal $x$, both including $N$ elements.
From (\ref{eq:dft}), the DFT of a vector $x$ can be written as the matrix multiplication  $y =\textbf{W}_\text{DFT} \, x$
with the weight matrix $\textbf{W}_\text{DFT}$ containing the given complex weights. By splitting $x$ into its real and imaginary parts,
$\Re(x)$ and $\Im(x)$, we can define the matrix multiplication with only real-valued variables:

\begin{equation}
\label{eq:real_dft}
    \begin{pmatrix}
        \Re(y) \\
        \Im(y)
    \end{pmatrix}
    = 
    \begin{pmatrix}
        \Re(\textbf{W}_\text{DFT}) & -\Im(\textbf{W}_\text{DFT}) \\
        \Im(\textbf{W}_\text{DFT}) & \Re(\textbf{W}_\text{DFT})
    \end{pmatrix}
    \begin{pmatrix}
        \Re(x) \\
        \Im(x)
    \end{pmatrix}
    \,.
\end{equation}

\begin{figure*}[h]
    \centering
    \includegraphics[width=\textwidth]{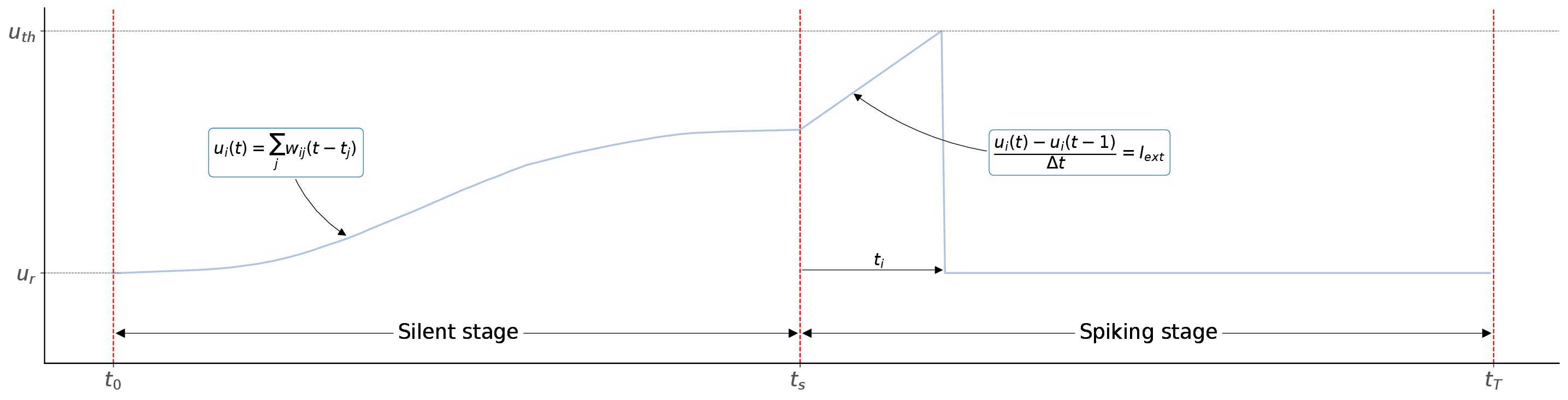}
    \caption{Sketch of the time-evolution of the time-coded S-DFT membrane voltage. In the first stage, pre-synaptic spikes charge the neuron to a voltage $u_i(t_T)$.
    In the second stage, the spiking of the neuron is forced by an external current $I_{ext}$. The voltage is reset at the end of the second stage.}
    \label{fig:ttfs_sdft}
\end{figure*}

For implementing precise matrix multiplications (\ref{eq:linear_combination}),
all spikes representing the input $x$ need to be causal,
i.e., neurons can only spike after all input spikes arrive.
To achieve this, the neuron operation is divided in two consecutive stages (see Figure \ref{fig:ttfs_sdft}).
During the first stage, the information contained in all input spikes is accumulated in the membrane voltage $u_i$
by setting a high threshold voltage $u_{th}$.
During the second stage, the neuron is charged with a fixed gradient that leads to an output spike at a time proportional to $u_i$.

\subsubsection*{Silent stage}

On an initial \textit{silent stage} during which the post-synaptic neuron $i$ does not spike, 
the membrane voltage of the neuron is modified following (\ref{eq:ttfs_membrane_update}) by the $N$ pre-synaptic spikes,
which arrive between the times $0$ and $t_s$.
Moreover, we add a constant bias $b_i$ to the neuron and substitute $t_j$ using (\ref{eq:ttfs_simplified_gamma}).
Hence, at time $t=t_s$,  (\ref{eq:ttfs_membrane_update}) results in
\begin{equation}
    \label{eq:snn_with_bias_long}
    u_i(t_s) = \sum w_{ij}\gamma x_j + \sum w_{ij} (t_s - \gamma x_\text{max}) + b_i\,.
\end{equation}
The bias $b_i$ is chosen as $b_i = - \sum w_{ij} (t_s - \gamma x_\text{max})$, so that the voltage $u_i(t_s)$ is directly proportional to $z_i$ in (\ref{eq:linear_combination}), 
\begin{equation}
    \label{eq:snn_with_bias}
    u_i(t_s) = \gamma \sum_j w_{ij} x_j \,.
\end{equation}
%
%
We choose $t_s$ to be the same for all neurons to keep the same time-to-value mapping (\ref{eq:ttfs_simplified_gamma}).
For the special case of representing a DFT without the offset bin ($i=0$), using a bias current in this stage is not required,
as the sum of all weights in (\ref{eq:dft}) is zero for all non-zero bins
$\{b_i = 0, \; \forall i \neq 0\}$.

Equation (\ref{eq:snn_with_bias}) is true only if neuron $i$ does not spike during the silent stage.
Thus, the membrane voltage $u_i$ cannot reach the threshold voltage $u_\text{th}$ during this stage.
We refer the reader to Appendix \ref{sec:appendix_charging} for a more detailed explanation.

To optimize the dynamic range of $u_i(t_s)$, $u_{th}$ is set to the boundary condition
\begin{equation}
    \label{eq:v_threshold}
    u_{th} = \max_{\forall i} \left\{u_{i,\text{max}}\right\} =  \max_{\forall i} \left\{\gamma \sum_j \left|w_{ij}\right| x_\text{max}\right\} \,.
\end{equation}
The maximum intensity $y_\text{max}$ that can be computed by the FT is given at the zero-frequency bin $i=0$ for a constant input $\{x_i = x_\text{max}, \; \forall i\}$.
The maximum intensity for a non-constant input is limited by half of $y_\text{max}$ for frequency bins $i\neq0$,
due to the symmetry property of the FT spectrum.
Therefore, the value of $u_{th}$ for the S-DFT can be further reduced and the calculations optimized by setting the threshold to
\begin{equation}
    \label{eq:v_threshold_dft}
    u_{th} = \dfrac{\gamma}{2} \sum_j w_{0j} x_\text{max} \,.
\end{equation}

\subsubsection*{Spiking stage}

To translate the voltage $u_i(t_s)$ into time-coded spikes,
neuron $i$ is charged on a \textit{spiking stage} by a constant current $I_\text{ext}$ from time $t_s$ until $t_{T}$ (see Figure \ref{fig:ttfs_sdft}).
The increase in the membrane voltage follows the linear function $\Delta u_i = \Delta t I_{ext}$.
Therefore, the neuron generates a spike at time $t_i$ when the membrane voltage $u_i$ reaches the threshold voltage $u_{th}$, which is determined by
\begin{equation}
    \label{eq:spiking_time}
    t_i-t_s = \dfrac{u_{th}-u_i(t_s)}{I_{ext}} \;.
\end{equation}
The interval $t_i-t_s$ is directly proportional to the output of the original function $y_i (x)$,
where the positive and negative values are represented by the first and second halves of the time range, respectively.
$t_i$ spans between $t_s$ and the total simulation time $t_T$.

The value of $I_{ext}$ is set to the minimum value that makes the neuron spike for all possible voltage values at $t_s$.
The most critical value is $u_i(t_s)_\text{min}$, which leads to  $min\{u_i(t_s)\} = - u_{th}$.
The external current $I_{ext}$ is then obtained from (\ref{eq:spiking_time}) as
\begin{equation}
    \label{eq:Iext_tuning}
    I_{ext} = \dfrac{2 u_{th}}{t_T-t_s} \,.
\end{equation}

\subsection{Network architecture}
\label{sec:architecture}

The proposed neuron model replicates a matrix-vector multiplication $z = \textbf{W} x$ using input and output neural layers connected by the weight matrix $\textbf{W}$. 
Thus, by applying feed-forward connections $\textbf{W}_l$ between the layers, it is possible to 
represent a sequence of $L$ matrix multiplications

\begin{equation}
    \label{eq:chain_multiplication}
    z = \textbf{W}_L \textbf{W}_{L-1} \ldots \textbf{W}_1 x \,.
\end{equation}

\begin{figure}[h]
    \centering
    \includegraphics[width=0.6\textwidth]{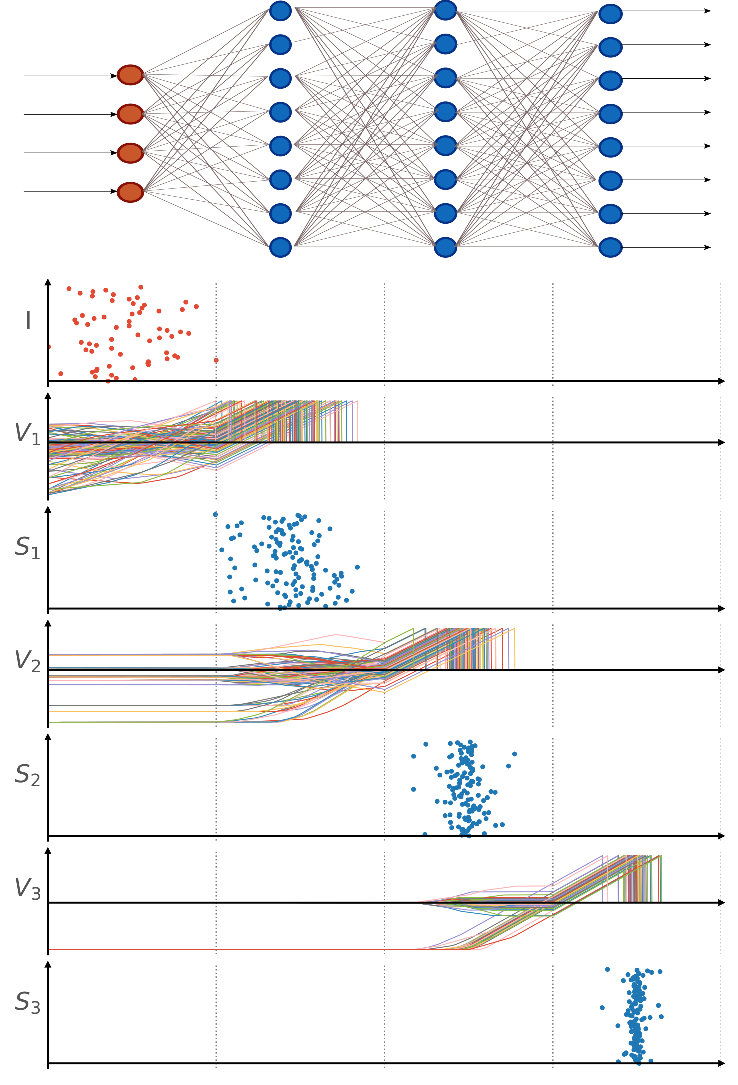}
    \caption{Representation of an SNN with an input layer and three hidden layers.
    The connections throughout the network are feed-forward,
    and the spiking stage of each layer $l$ overlaps with the silent stage of layer $l+1$.
    During the silent stage, the layer charges on the basis of spikes from the previous layer,
    and during the spiking stage the layer charges at a constant rate until reaching $u_{th}$.
    Therefore, neurons only produce spikes during the spiking stage.}
    \label{fig:architecture_drawing}
\end{figure}

To transmit spikes between layers, the spiking stage of a pre-synaptic layer overlaps with the silent stage of the corresponding post-synaptic layer connected right after.
This is due to the nature of the SNN, as neurons collect information from pre-synaptic connections during the silent stage, 
and then generate information during the spiking stage.
This means that for an SNN with $L$ layers, the total number of stages over time is $L+1$,
and a given layer $l$ receives spikes during the \textit{l-th} stage and generates spikes during the \textit{l+1-th} stage.
Figure \ref{fig:architecture_drawing} depicts a representation of the architecture and the overlap of the different stages.

Due to the nature of the SNN, the network does not need to wait for an input vector to be processed before feeding a new vector.
The frame rate of the network is only determined by the maximum time required for a single stage to be computed,
i.e., $ T_f = {t_T}_\text{max}$, where $T_f$ is the period between consecutive frames.
The latency of the network to process a single frame is the sum of the time required for processing spikes on each layer,
i.e., $\tau_f = \sum_l^{L} \tau_l$, where $\tau_f$ represents the time delay for processing a frame, and $\tau_l$ is the execution time of layer $l$.

We exploited this property for reproducing the linear combinations of an FT that operates in more than one dimension,
which is explained in more detail in \cite{lopez2021spiking}.
\\

\subsubsection{Spiking fast Fourier Transformation}

Here, we show an alternative architecture that takes advantage of the chained matrix multiplication (\ref{eq:chain_multiplication})
and reproduces the structure of an FFT algorithm.
By factorizing the DFT matrix into linear combinations of sparse matrices $\textbf{S}$,
the number of computations and connections can be reduced and an optimized FT can be obtained,
\begin{align}
    y = \textbf{W}_\text{DFT} \cdot x
              = \textbf{S}_L \textbf{S}_{L-1} \ldots \textbf{S}_1 \cdot x \;.
\end{align} 

The most common FFT algorithms exploit the symmetry properties of the complex weights of the DFT and recursively map the DFT to smaller DFTs,
being the smallest DFT called butterfly matrix \cite{FFTreview1990}.
This recursive mapping splits the FT computation into several consecutive sparse matrix multiplications that reduce the order of complexity. 
The number of matrix multiplications depends on the number of data points $N$ and the size of the butterfly matrix.
Here, we use the decimation-in-frequency Radix-4 butterfly matrix to build the sparse matrix representations $\textbf{S}$.
As a result, the number of stages or layers $L$ is determined by $L = \log_4 (N)$ \cite{FFTcomparison1999}.
A detailed explanation of the Radix-4 algorithm, its butterfly matrix, and complex weights can be found in \cite{FFTreview1990} and \cite{FFTcomparison1999}.

The sparse matrices $\textbf{S}$ consist of multiplications of complex 4-by-4 radix-4 butterlfy matrices
\begin{align}
    \textbf{B}_\text{4x4} =
\begin{pmatrix}
        1 & 1 & 1 & 1\\
        1 & -i & -1 & i\\
        1 & -1 & 1 & -1\\
        1 & i & -1 & -i
\end{pmatrix},
\end{align}
and complex diagonal weight matrices 
\begin{align}
 \textbf{W}_\text{4x4} =
 \footnotesize\begin{pmatrix}
        W^0_N & 0 & 0 & 0\\
        0 &W^k_N & 0 & 0\\
        0 & 0 &W^{2k}_N & 0\\
        0 & 0 & 0 &W^{3k}_N
\end{pmatrix},
\end{align}
with $W^k_N = e^{i 2 \pi k / N }$.
As the neuron model only works with real values, the complex matrix $\textbf{B}_\text{4x4} \cdot \textbf{W}_\text{4x4}$ is transformed into 
a real-valued 8-by-8 radix-4 matrix $\textbf{W}_\text{8x8} \cdot \textbf{B}_\text{8x8}$, where half of the connections represent the imaginary components.
By using the same rephrasing as in (\ref{eq:real_dft}), the transformation yields the result

\begin{align}
\textbf{W}_\text{8x8} \cdot \textbf{B}_\text{8x8} = \footnotesize\begin{pmatrix}
   \Re(\textbf{W}_\text{4x4}) & - \Im(\textbf{W}_\text{4x4})\\
   \Im(\textbf{W}_\text{4x4}) & \Re(\textbf{W}_\text{4x4})
   \end{pmatrix}
   \cdot 
   \begin{pmatrix}
   \Re(\textbf{B}_\text{4x4}) & - \Im(\textbf{B}_\text{4x4})\\
   \Im(\textbf{B}_\text{4x4}) & \Re(\textbf{B}_\text{4x4})
   \end{pmatrix}
   \,.
\end{align}

Instead of using an all-to-all connection layout as in the DFT matrix, only up to eight connections per neuron are needed.
As a spike generated in one neuron has to be distributed to all neurons connected to its output, the number of spike operations (ops) is given by the number of connections per neuron
$N_\text{conn}$, the total number of neurons $N_\text{neurons}$, and the number of output spikes $N_\text{output}$. 
Assuming the same number of connections per neuron, the number of spike ops is given by
\begin{equation}
    \label{eq:spike_ops}
    N_\text{spike ops.} = N_\text{conn} \cdot N_\text{neurons} + N_\text{output} \,.
\end{equation}
Based on (\ref{eq:spike_ops}), the number of spike ops of the S-DFT is determined by
\begin{align}
    N_\text{S-DFT} = 2 N \cdot 2 N + 2N \,,
\end{align}
whereas the spiking FFT (S-FFT) version requires
\begin{align}
    N_\text{S-FFT} = 8 \cdot 2 N \cdot \log_4(N) + 2N
\end{align}
spike ops. In addition, the S-FFT requires $\log_4(N)$  layers and $2 N$  neurons per layer, including both real and imaginary values.
In Table \ref{tab:s-ft_summary}, we compare the reduction of connections in the S-FFT network with the reduced number of neurons in the S-DFT network.

Thus, the S-FFT network reduces the number of spike operations by increasing the total number of neurons. We evaluated the benefits of
using fewer spike operations in Section \ref{sec:performance} by comparing neuromorphic implementations of both the S-DFT and S-FFT.

\begin{table}[h]
    \centering
    \renewcommand{\arraystretch}{1.3}
    \setlength{\tabcolsep}{12pt}
    \caption{Summary of the network parameters for an S-FFT and S-DFT that process an input chirp with $N$ samples,
    and has a delay $\tau_l$ per stage.}
    \begin{tabular}{ccc}
        \hline
        \textbf{Parameter}  & \textbf{S-FFT}                    & \textbf{S-DFT} \\ 
        \hline
        Nº layers     & $\log_4(N)$                         & $1$ \\
        Nº neurons    & $2 N \cdot \log_4(N) $              & $2 N$ \\
        Nº spike ops. & $8 \cdot 2 N \cdot \log_4(N) +2N $  & $N \cdot 2 N + 2 N$ \\
        $T_f$         & $2 \cdot \tau_l $                   & $2 \cdot \tau_l $ \\
        $\tau_f$      & $\tau_l \cdot (\log_4(N)+1) $       & $2 \cdot \tau_l $ \\
        \hline
    \end{tabular}

    \label{tab:s-ft_summary}
\end{table}

\subsection{Neuromorphic hardware implementation}

To assess the feasibility of accelerating the network in neuromorphic hardware and evaluating its performance, the proposed SNN has been implemented on Intel's digital research chip Loihi \cite{davies2018loihi}.  
The chip consists of 128 neuromorphic cores, and each core can integrate 1024 spiking neural units, called compartments. 
Three embedded Intel Lakemont x86 processor cores manage the neuromorphic cores and control the spike traffic that is directed in and out of the chip.
The chip is a fully-digital many-core mesh that implements the current-based leaky integrate-and-fire (CUBA LIF) neuron model.
The compartments are implemented as homogeneous groups that share the basic structure and parameters. Connections are configured similarly: synapses connecting two populations of neurons show the same functional behavior and only differ in their weights.

To implement the proposed neuron model (\ref{eq:ttfs_membrane_update}) on Loihi, 
the parameters of the inherent CUBA LIF model have to be adjusted. The voltage membrane of the standard implementation
follows the differential equation
\begin{align}
    \frac{du_i}{dt}(t) = -\frac{1}{\tau_u} u_i(t) + I^\text{syn}_i(t) + b_i \,,
\end{align}
with a voltage time constant $\tau_v$, bias $b_i$, and synaptic response current $I^\text{syn}_i(t)$ of neuron $i$. 
The synaptic current depends on the incoming spike $s_j = \delta(t-t_j)$ as
\begin{align}
    I^\text{syn}_i(t) = \sum_j w_{ij} \frac{1}{\tau_I} e^{-t/\tau_I} H(t) \delta(t-t_j)   \,,   
\end{align}
with a current time constant $\tau_I$, Heaviside function $H(t)$ and weights $w_{ij}$.
The S-FT dynamics (\ref{eq:ttfs_membrane_variation}) have no leakage, $1/\tau_u = 0$,
and no synaptic decay, $(1/\tau_I) e^{-t/\tau_I} = 1$.
The synapses store and accumulate the weights of the incoming spikes without decay
and drive the voltage over time. Up to this point, the calculations of the neuron model are 
limited only by the 8-bit precision of weights on the Loihi chip.
Since we further rely on the silent stage for our neuron calculations,
the threshold has to be set accordingly. 
The accuracy of the calculations is further constrained by the existence of an upper limit of the voltage threshold in Loihi.

For the transition between the silent and spiking stages, the synaptic current is reset, and a fixed synaptic bias is introduced to each neuron,
inducing a constant increase in the membrane potentials, that eventually reach the voltage threshold and generate a single spike per neuron.

\begin{figure}[h]
    \centering
    \includegraphics[width=0.45\textwidth]{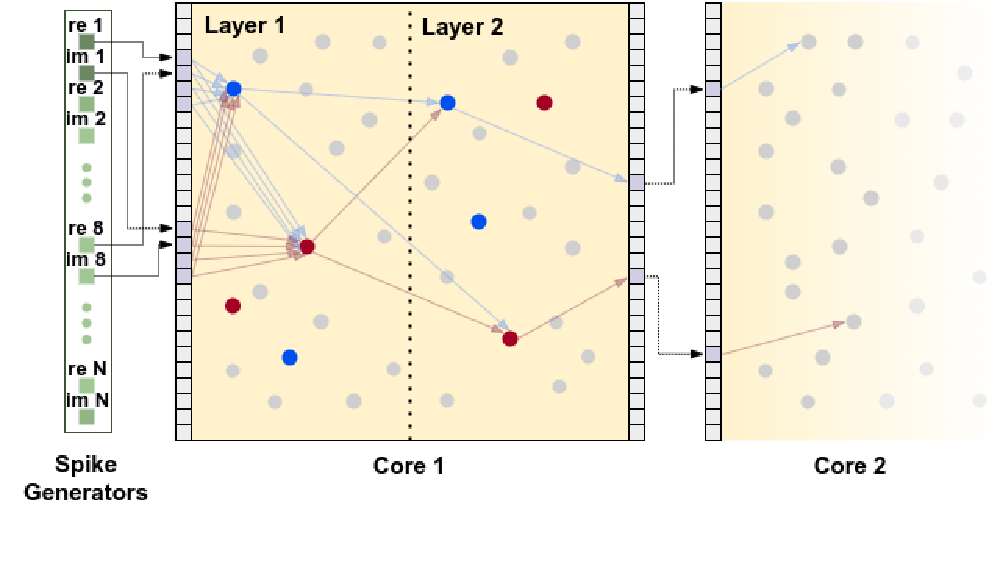}
    \caption{
    Representation of the S-FFT distribution in the Loihi chip. In green, the spike generators feeding the information from the $N$ input samples.
    In blue, four arbitrary neurons representing the real component of two values in the first layer and two values in the second layer, respectively.
    In red, four neurons representing the imaginary components of the aforementioned four values.
    Due to hardware limitations, some connections are implemented in the same core. This is the case of the ones between layers 1 and 2 in the figure.
    Other connections, like the output of layer 2, are routed between different cores.
    }
    \label{fig:core_distribution}
\end{figure}

The two architectures introduced in Section \ref{sec:architecture} take encoded radar data as input.
For the S-DFT, each input node is connected to every neuron and weighted with the corresponding DFT coefficient.
For the S-FFT, the connection matrix is sparse, with each neuron connected to eight inputs (four real and four imaginary).
Figure \ref{fig:core_distribution} illustrates the distribution of the S-FFT in the chip.
The input data affect two key configuration features of the network:
First, it determines the network size, as the number of samples in the radar data dictates the number of spike generators and the number of neurons in each layer;
second, it affects the run time, as the process of encoding and feeding the data to the network with a designated resolution takes a fixed number of time steps $n_T$.
This determines the duration of the silent and spiking stages, as the membrane dynamics of each stage (\ref{eq:snn_with_bias_long}) are simulated on discrete time over $n_T$ steps.
Increasing the number of time steps will improve the resolution of the network at the expense of a \mbox{prolonged execution time.}

The accumulation of incoming weighted spikes from all input generators requires a high value range from Loihi's network parameters.
The membrane potential of each neuron is given a central role in these calculations. It takes values in the range $[-2^{23}, 2^{23}]$, with $2^{23}-2^6$ being the maximum threshold.
This threshold cannot be reached during the silent stage.
The current induced by each spike is also limited to a fixed minimum value of $2^6$.
The synaptic weights $w$ can be excitatory and inhibitory, and they are based on the FT equations.
They are implemented in Loihi according to
\begin{equation}
    w = m * 2^{exp} \,,
\end{equation}
where $m$ represents the mantissa, and $exp$ is the exponent. They can take integer values in the range $[-2^8,\; 2^8-1]$ and $[-8, 7]$, respectively.
Moreover, should the full range of $m$ be used, it can only take even values.

The aforementioned variable ranges introduce a bottleneck in the implementation, limiting the precision of the network,
which becomes more relevant as the number of input samples increases.

\section{Validation experiments}
\label{sec:experiments}

To evaluate the performance of the proposed S-FT in realistic conditions,
we collected data in real-world scenarios using a radar sensor, which is prompt to clutter perturbations.
We designed tailored experiments in which the radar sensor was exposed to challenging corner cases, and compared the spiking result with the output from a dedicated FFT accelerator \cite{iscaslowfootprintaccelhg}.

The code and data for running the experiments are available at an open-source repository\footnote{\url{https://github.com/KI-ASIC-TUM/time-coded-SFT}}.

\subsection{Radar dataset}\label{sec:dataset}
We have collected raw radar data using a commercial \mbox{77-GHz} frequency modulated continuous wave (FMCW) radar (AWR1642Boost-ODS) from Texas Instruments. 

\begin{figure}[h]
    \centering
    \includegraphics[width=0.45\textwidth]{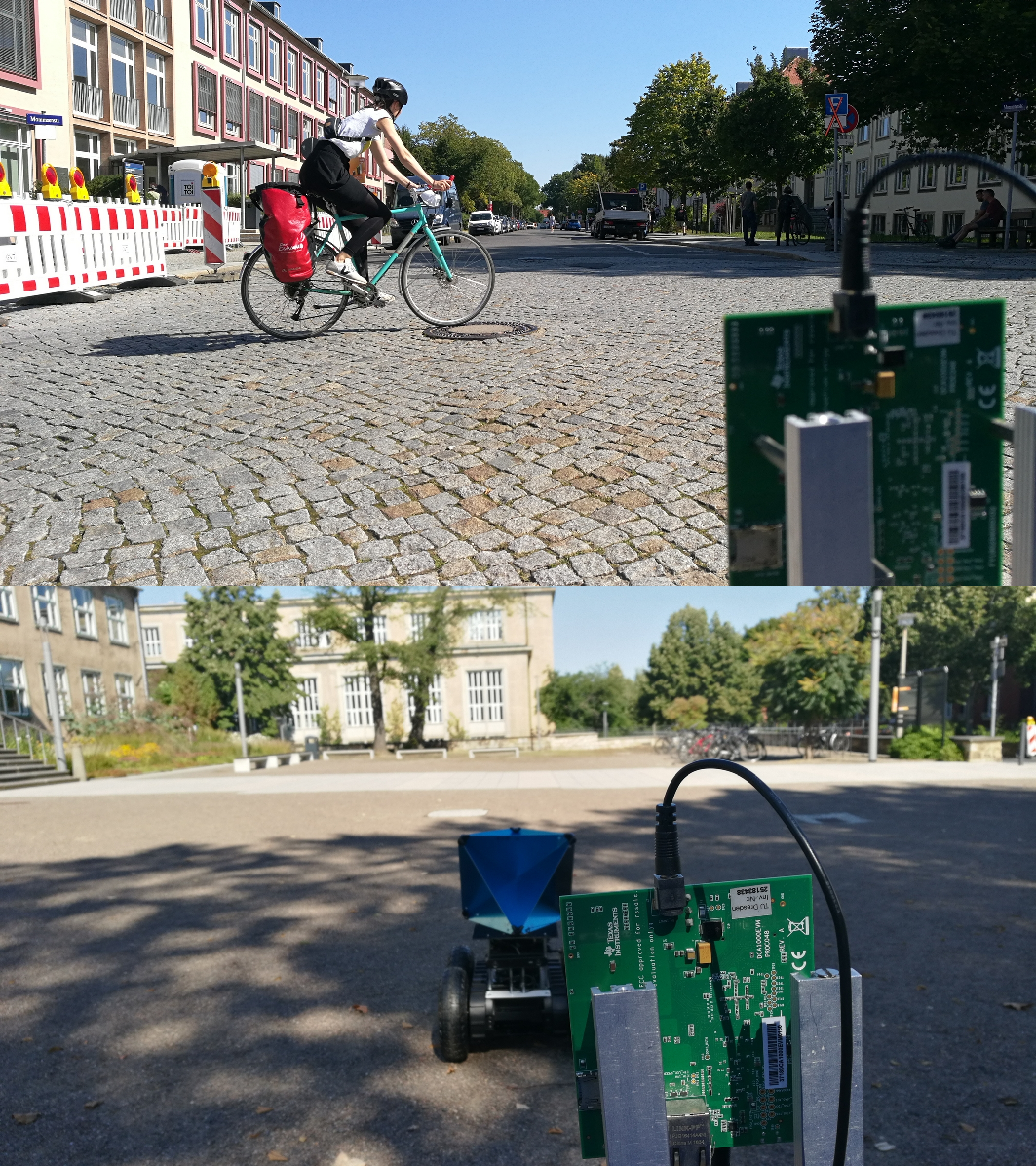}
    \caption{
    Images with the two setups used for experimental radar recordings. On the top image, the radar faces objects with different radar cross-section.
    On the bottom image, the radar faces a mobile robot carrying a corner detector.
    }
    \label{fig:scenarios}
\end{figure}

Two representative setups were considered during the recording.
As shown in Figure \ref{fig:scenarios}, the radar sensor is statically erected in an empty yard with a height of \SI{1}{\meter}.
In the first setup, the radar faces a crossing, where different objects with varied radar cross-section, e.g., pedestrians, cyclists and cars, are captured.
In the second setup, a mobile robot delivers a radar corner reflector that returns high-intensity radar echoes in the front.
The robot moves within the range of \SI{22}{\meter} slowly (radial velocity between \SI{-1}{\meter/\second} and \SI{1}{\meter/\second}).

To ensure a long length of FFT benchmarking up to 1024 points, the radar configuration is well-designed, as described in Table \ref{tab:radar setting}.
\begin{table}[h]
    \centering
    \renewcommand{\arraystretch}{1.3}
    \setlength{\tabcolsep}{12pt}
    \caption{Summary of radar configuration.}
    \begin{tabular}{ll}
        \hline
        \textbf{Parameter}                      & \textbf{Value}     \\ \hline
        Bandwidth (\si{MHz})                    & $1535$    \\
        Sampling frequency (\si{MHz})           & $5$       \\
        Chirps per frame                        & $128$     \\
        Chirp time (\si{\us})                   & $230$     \\
        Range Max. (\si{m})                     & $56.2$    \\
        Range Res. (\si{m})                     & $0.1 $    \\
        Velocity Max. (\si{\meter/\second})     & $2.0$     \\
        Velocity Res. (\si{\meter/\second})     & $0.06$    \\
        \hline
    \end{tabular}
    \label{tab:radar setting}
\end{table}

\subsection{FFT accelerator used for comparison}
\label{sec:fftaccel}

\begin{figure}[h]
    \centering
    \includegraphics[width=0.45\textwidth]{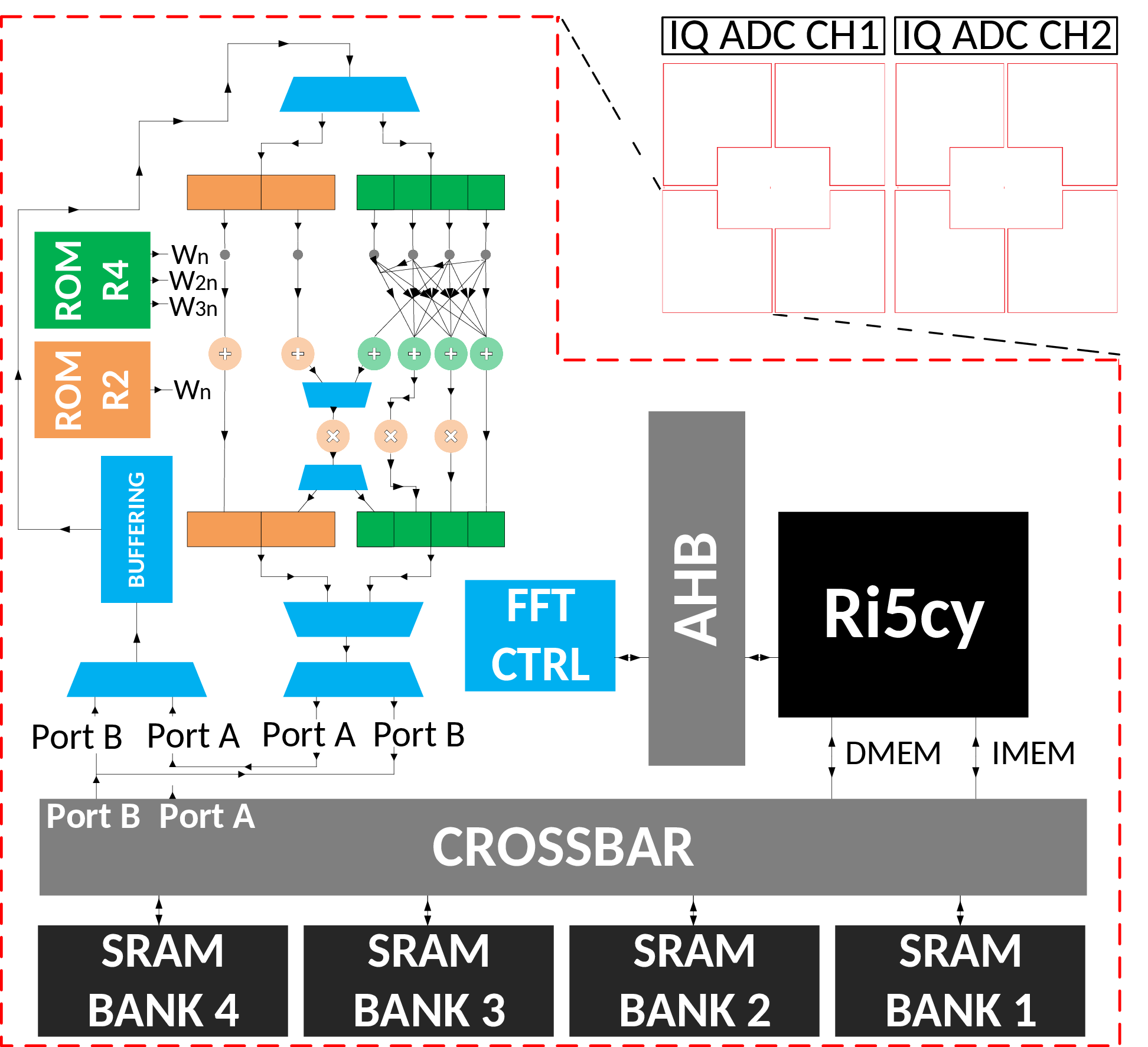}
    \caption{
    Block diagram of the FFT accelerator used as reference (\cite[Figure~1]{iscaslowfootprintaccelhg}).
    }
    \label{fig:accelFFT}
\end{figure}

The FFT accelerator used to compare the performance of the S-FT is a 22FDX memory-based approach described in \cite{iscaslowfootprintaccelhg} and shown in Figure \ref{fig:accelFFT}, which is highly-optimized for the radar processing chain in terms of latency and area with a \mbox{dual-radix} butterfly. The accelerator uses a Radix-4 butterfly for the Range-FFT at a resolution of 16 bits, whereas it uses a Radix-2 buttefly for the Doppler and Angle FFT at 32 bits.
\\

Despite this accelerator having a dual-radix butterfly, and allowing the utilization of either a Radix-2 (32 bits) or a Radix-4 (16 bits) mode, the latter is used in this study to employ a more comparable counterpart for the S-DFT. The accelerator maintains a high throughput by aligning \mbox{non-consecutive} operators via a buffering stage that ensures a single result is always written in every clock cycle.

\subsection{Results}

We have tested the proposed SNN on the dataset introduced in section \ref{sec:dataset}.
The purpose of this experiment is to validate the SNN by running it through different scenarios and evaluating the error for each one of them. 

We have computed the root mean square error (RMSE) between the result of the scientific library NumPy\footnote{\textit{https://numpy.org/doc/stable/reference/routines.fft.html}} on a general-purpose computer, and the results of the S-FT and the FFT accelerator introduced in Section \ref{sec:fftaccel}, respectively, using the equation
%
\begin{equation}
    \label{eq:rmse}
    RMSE(X, Y) = \sqrt{\dfrac{\sum_{i=0}^{N-1}(X_i - Y_i)^2}{N} }\,,
\end{equation}
where $X$ and $Y$ represent two signals of size $N$.

We have also generated the plots of the SNN and FFT accelerator outputs for specific configurations to visually compare both signals and spot local deviations in specific bins or scenarios. 



We have analyzed the accuracy of the S-FT in four static scenarios that include typical challenging situations for radar sensors:

\begin{enumerate}
    \item One strong reflection close to the sensor and one weak reflection far away
    \item A weak reflection far away
    \item Two reflections close to each other
    \item Multiple reflections
\end{enumerate}

\begin{figure*}[h]
    
    \begin{subfigure}{.24\textwidth}
        \centering
        \includegraphics[width=1\linewidth]{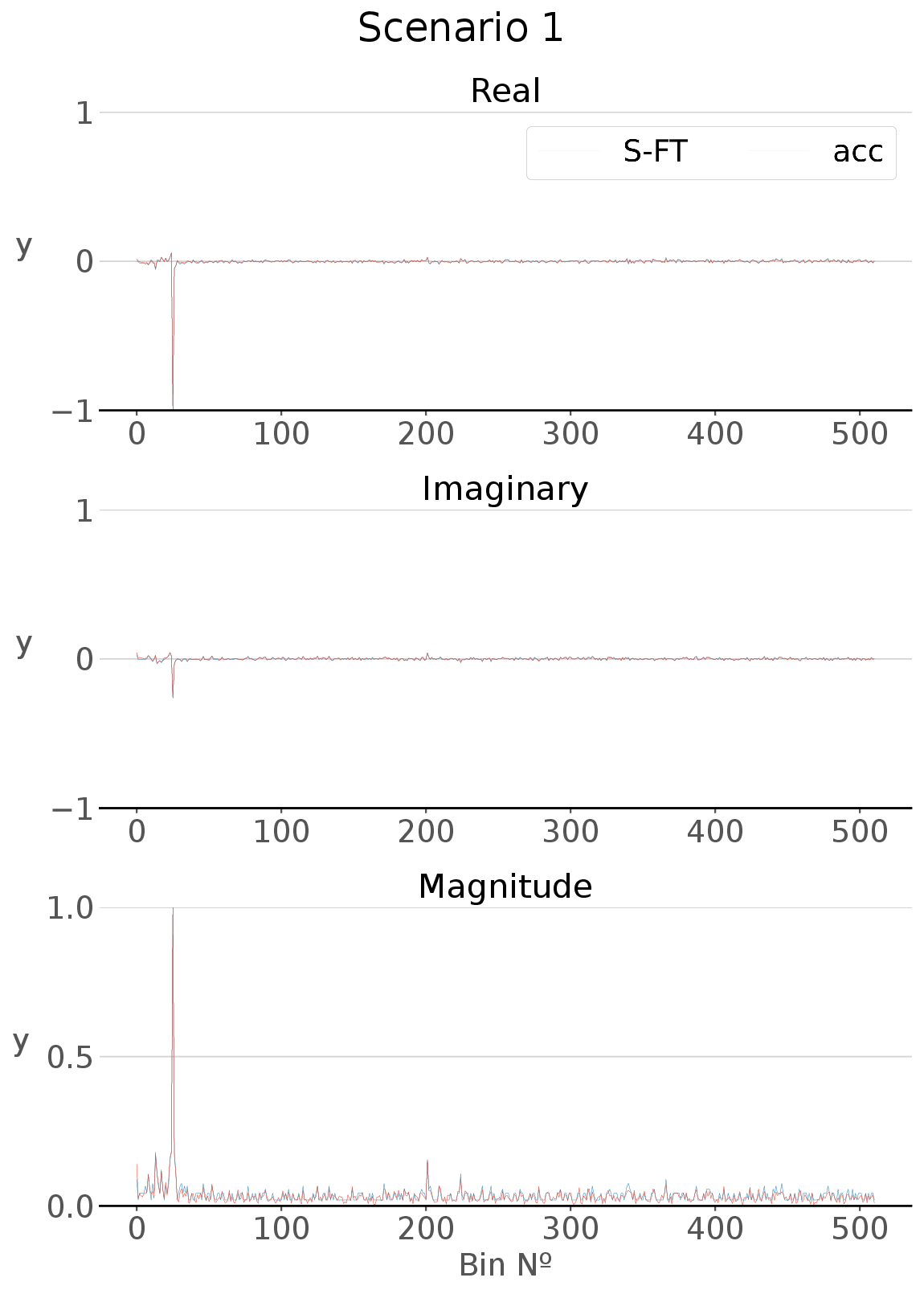}
    \end{subfigure}
    \begin{subfigure}{.24\textwidth}
        \centering
        \includegraphics[width=1\linewidth]{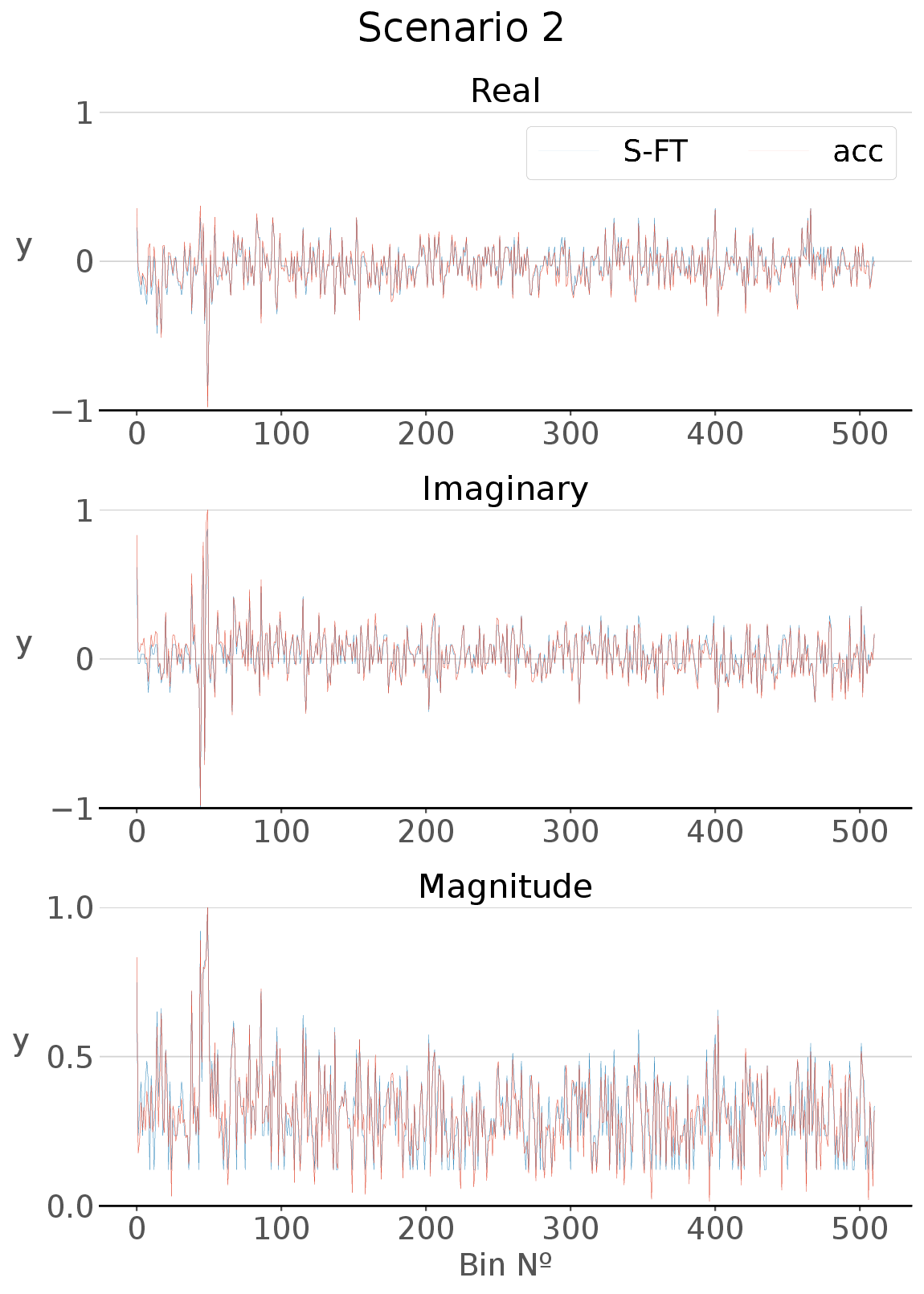}
    \end{subfigure}
    \begin{subfigure}{.24\textwidth}
        \centering
        \includegraphics[width=1\linewidth]{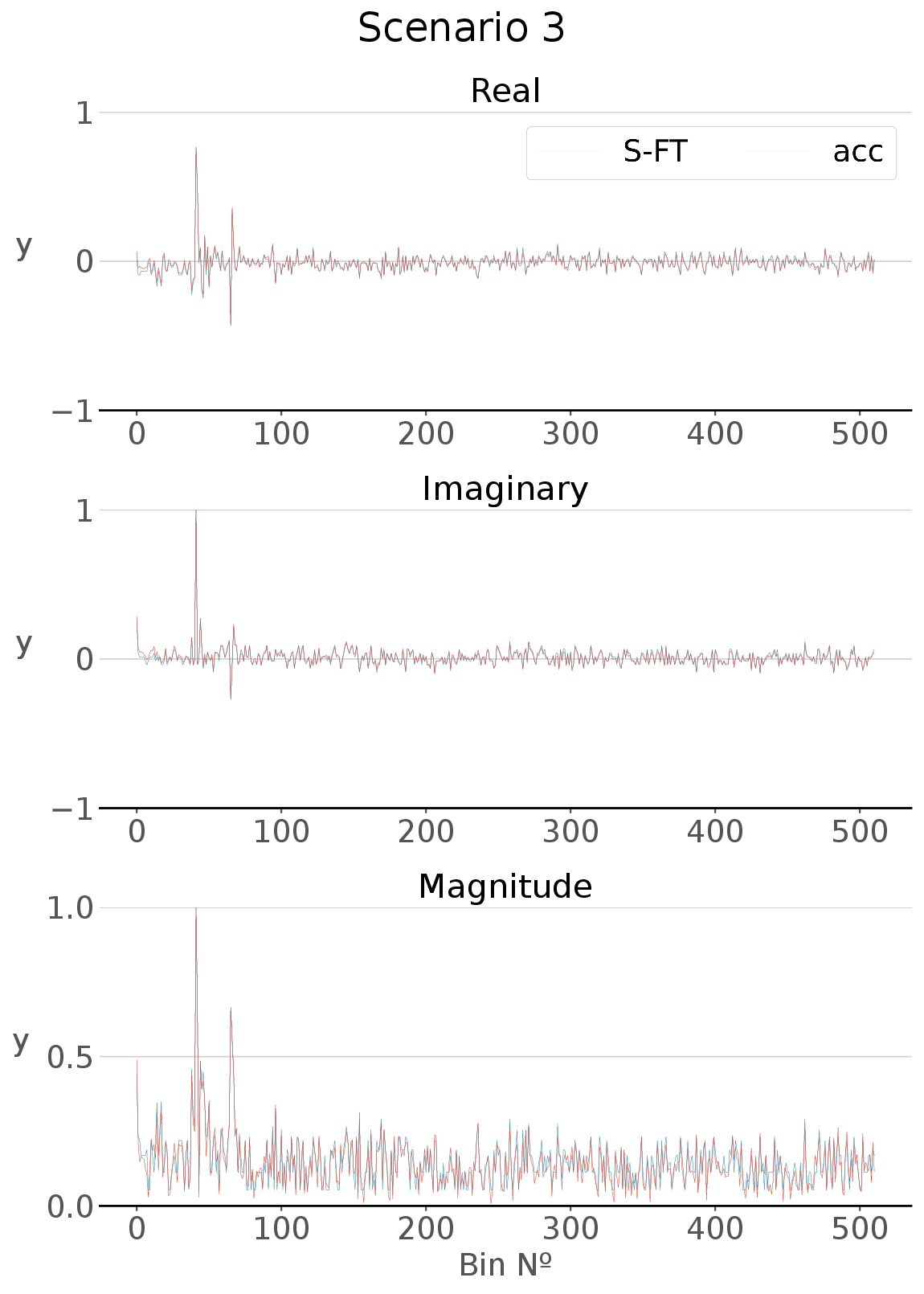}
    \end{subfigure}
    \begin{subfigure}{.24\textwidth}
        \centering
        \includegraphics[width=1\linewidth]{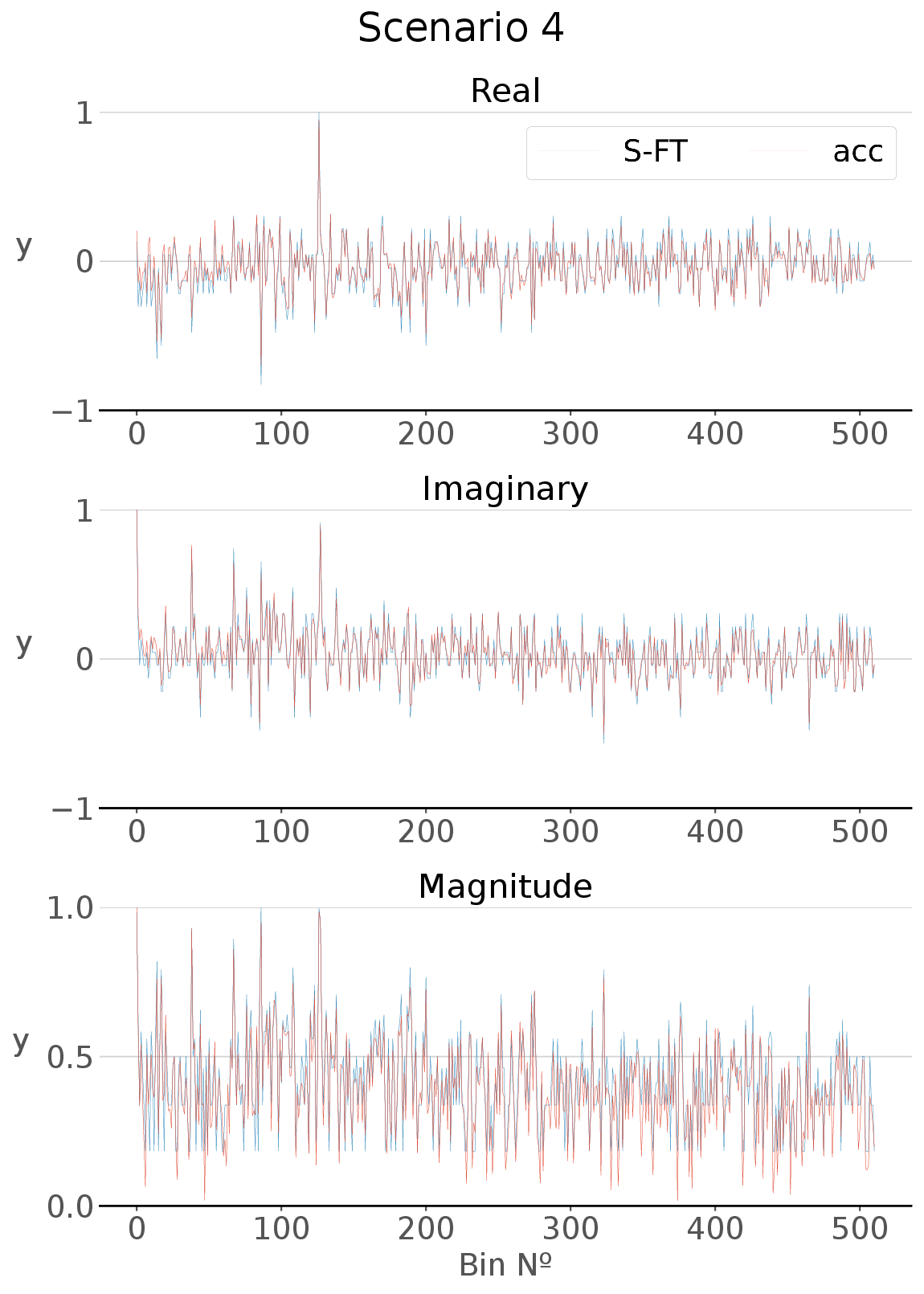}
    \end{subfigure}
    \caption{Error of the S-FFT for four different corner cases. In blue, the output of the S-FFT implemented in Loihi.
    In red, the reference signal provided by the FFT accelerator introduced in Section \ref{sec:fftaccel}.
    The experiments have been conducted for a bin size of $1024$ and for $257$ time steps per SNN simulation stage.
    A signal with only real values is used as input, resulting in a symmetric FT. 
    Thus, only the positive side of the spectrum is displayed.
    In addition, the offset bin has been removed, as its information is irrelevant for the range dimension of a radar.
    The results have been normalized between $-1$ and $1$ for the real and imaginary terms.
    The magnitude plot depicts the logarithm of the output value normalized between $0$ and $1$.}
    \label{fig:error_multi}
\end{figure*}

Figure \ref{fig:error_multi} shows the output of the S-FFT and the reference FFT for each of the aforementioned cases.
In all cases, we have run the network with a stage time of 257 time steps, and for a data length of 1024 samples.
Table \ref{tab:rmse_summary} depicts the RMSE of the S-DFT, S-FFT, and the reference FFT accelerator
after correcting the offset and normalizing the output to the same range.
The table shows an error of \textit{0.041} in the worst-case scenario. 
In addition, the error is distributed homogeneously over the transform, thereby preserving the information contained in the intensity difference between bins.

\begin{figure}[h]
    \centering
    \includegraphics[width=0.45\textwidth]{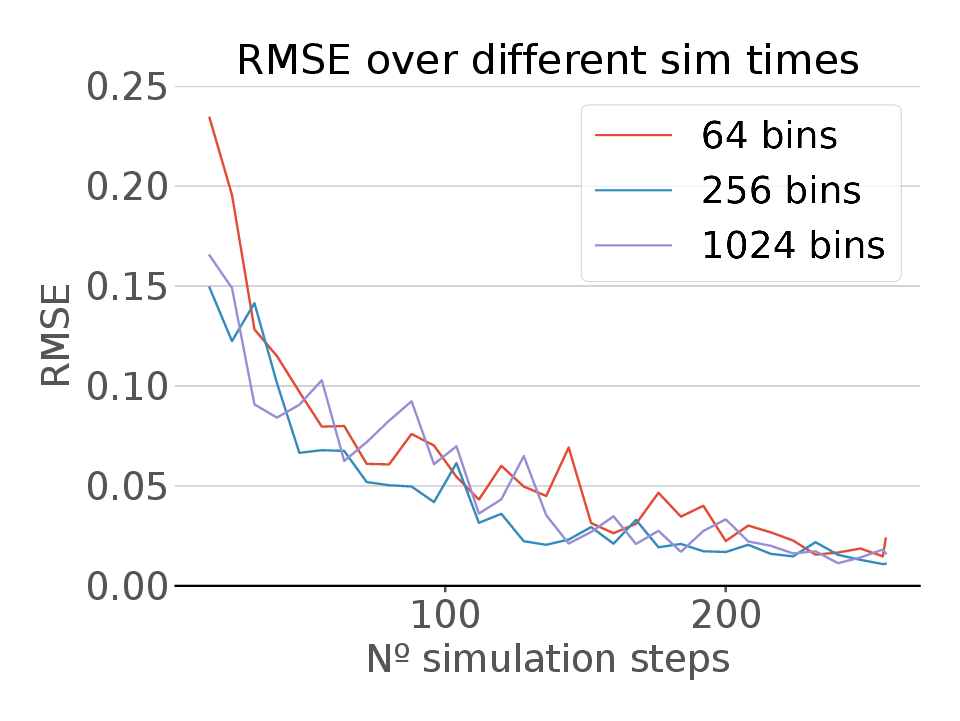}
    \caption{
    RMSE of the S-FFT in Loihi for different simulation times when compared to the reference FFT algorithm.
    The \textit{x} axis indicates the amount of simulation steps on each of the SNN stages.
    The experiment has been performed for FT sizes of 64, 256, and 1024 bins, respectively.
    }
    \label{fig:rmse_plot_multi}
\end{figure}

Figure \ref{fig:rmse_plot_multi} shows how the number of time steps affects the accuracy of the S-FFT.
We have conducted the experiment for three different bin configurations compatible with the \mbox{radix-4} architecture
and calculated the RMSE over the four static scenarios.

\begin{table}[h]
    \centering
    \renewcommand{\arraystretch}{1.3}
    \setlength{\tabcolsep}{12pt}
    \caption{RMSE of the S-DFT and S-FFT compared to the output from \textit{NumPy} for $n_\text{bins}=256$ and $t_\text{s}=256$.}
    \begin{tabular}{ccccc}
        \hline
        Architecture      & S1    & S2    & S3    & S4    \\ \hline
        \textbf{S-DFT} & 0.004 & 0.041 & 0.009 & 0.030 \\
        \textbf{S-FFT} & 0.006 & 0.026 & 0.007 & 0.028 \\ 
        \textbf{Accelerator} & 0.0005 & 0.0005 & 0.0005 & 0.0005 \\ 
        \hline
    \end{tabular}
    \label{tab:rmse_summary}
\end{table}

\begin{figure}[h]
    \centering
    \includegraphics[width=0.45\textwidth]{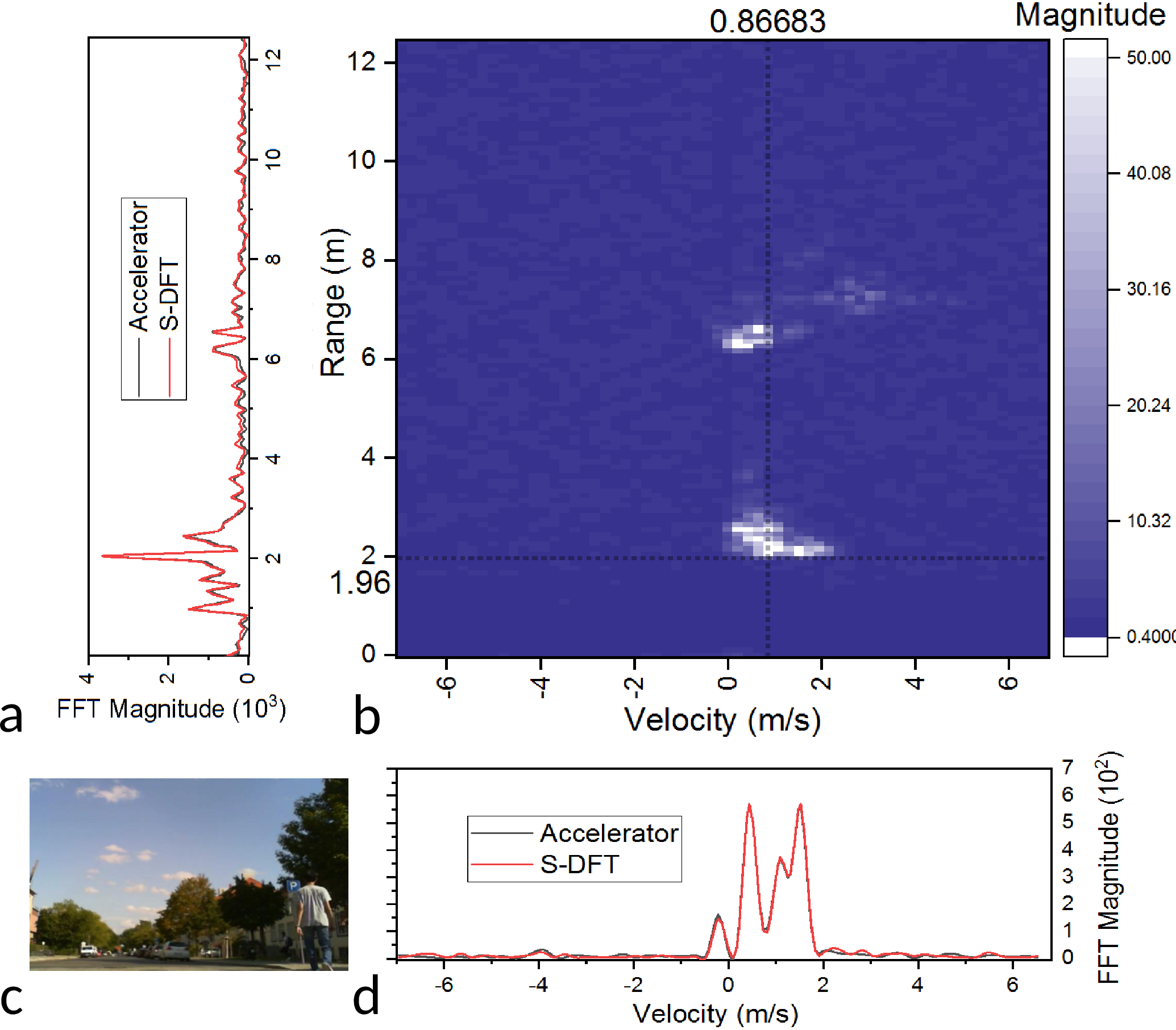}
    \caption{
    Conventional and spike-based range (a) and Doppler (d) FFT for the Range-Doppler map (b) of the scenario (c).
    }
    \label{fig:rdmap}
\end{figure}

In addition, we have assessed the S-FFT on a scenario with dynamic objects that introduce changes in the Doppler dimension of the radar signal,
i.e., by performing an FT over consecutive chirps of a frame the velocity of the different objects can be obtained.
Figure \ref{fig:rdmap} depicts the performance of the S-FFT on the range dimension for one of the input chirps, as well as the Doppler dimension for a specific range bin.
The S-FFT follows the reference with minor variations outside the main peaks, as shown by the cross-section FT in Figures \ref{fig:rdmap}a and \ref{fig:rdmap}d.

\subsection{Computational performance}
\label{sec:performance}

We provide in this section a notion of the main aspects of the computational performance of the proposed SNN when implemented in the neuromorphic chip Loihi (See Table \ref{tab:FOMffts}).



The values offered for the Loihi implementation are estimated with the results of the performance experiments conducted in \cite{davies2018loihi}
and combined with the S-DFT and S-FFT computational parameters from Table \ref{tab:radar setting}.
The energy and execution time per frame are calculated using
\begin{equation}
    \label{eq:loihi_energy}
    E = n_\text{spikes} \cdot 23.6 \text{pJ} + n_\text{steps} \cdot n_\text{neurons} \cdot 52 \text{pJ} \,
\end{equation}
and
\begin{equation}
    \label{eq:loihi_time}
    \tau_f = n_\text{spikes} \cdot 3.5 \text{ns} + n_\text{steps} \cdot n_\text{neurons} \cdot 8.4 \text{ns} \,.
\end{equation}

As the execution of SNNs can be highly parallelized in neuromorphic hardware,
we reduce (\ref{eq:loihi_time}) to the time required to process all spikes and neuron updates in a single neuromorphic core.
In the case of Loihi, the execution is distributed among 128 neural cores, resulting in 16 and 80 neurons per core for an S-DFT and S-FFT with 1024 bins, respectively.

In addition, the multilayer architecture of the S-FFT allows processing the incoming chirp $n$ once the previous chirp $n-1$ has already passed the first layer.
This implies that the total energy consumed per chirp by the neuron updates through the total time $\tau_f$ is divided by the total number of chirps that can be simultaneously processed.
In other words, the S-FFT uses only $1/L$ of its neurons at every processing stage while the remaining neurons can handle other chirps in the meantime.
\\

\begin{table}[h]
    \centering
    \renewcommand{\arraystretch}{1.3}
    \setlength{\tabcolsep}{12pt}
    \caption{Figure of Merit for the proposed architectures in the neuromorphic chip Loihi
    for processing data streams of 1024 samples and 75 simulation steps per processing stage.}
    \begin{tabular}{ccc}
    \hline
        \textbf{Parameter}              & \textbf{S-DFT}    & \textbf{S-FFT}\\ \hline
        $n_\text{neurons}$              & 2048              & 10240         \\
        $n_\text{spike ops.}$ (thsnd.)  & 2100              & 84            \\
        $E$ ($\mu J$)                   & 65.5              & 49.9          \\
        $T_f$ ($\mu s$)                 & 77.6              & 105           \\
        $\tau_f$ ($\mu s$)              & 77.6              & 315           \\
        $P$ (mW)                        & 844               & 158           \\
        \hline
    \end{tabular}
    \label{tab:FOMffts}
\end{table}

\section{Discussion}

We have evaluated the proposed SNN on real-world radar data.
We tested the S-DFT and S-FFT architectures on five different scenarios that pose different challenges,
e.g., target intensity, distances between targets, scenario complexity, and dynamic properties.
In all cases, the proposed SNN provides a low-error output and can reproduce the result of the reference algorithm for all FT bins.
The error is caused by the loss of dynamic range when converting to spikes and 
the low resolution of Loihi's parameters, such as the weights and membrane voltage.
The low error holds when the total number of output bins changes, and it improves when the number of simulation time steps is increased (see Figure \ref{fig:rmse_plot_multi}).

We have also estimated the computational performance of both architectures in Loihi.
Table \ref{tab:FOMffts} shows that the S-DFT outperforms the S-FFT in terms of execution time,
due to the larger number of neurons in the latter.
Regarding the total energy required to process one chirp, the S-FFT offers better performance,
because each neuron  of this network is only used during a small fraction of the runtime,
i.e., during its corresponding silent and spiking stage (see \mbox{Figure \ref{fig:architecture_drawing})}.

State-of-the-art FFT accelerators include memory-based DSPs in \cite{ref31} and \cite{iscaslowfootprintaccelhg}, the latter being a very attractive solution for multi-core DSPs in MIMO FMCW radars, as its footprint is minimal; or contributions for high-performance applications such as the LTE communication chip in \cite{matrixfftref2}. The energy consumption of these chips for the FT parameters used in Section \ref{sec:performance} is  $484$, $56.3$, and $126$ nJ, respectively,
whereas the execution time is $2.81$, $8.8$, and \mbox{$1.38$ $\mu s$}, respectively. It must also be noted that the matrix-based FFT accelerator described in \cite{matrixfftref2} is optimized for high performance, which implies that the latency benefits are only evident for FFT lengths larger than 1024 points.
Such lengths have not been covered by this work, as they were not supported by the other FT accelerators and fall outside of the FMCW radar scope.
Table \ref{tab:FOMffts} shows that the performance of the 
state-of-the-art FT accelerators is  $9 \text{ - } 76$ times more efficient than that of the neuromorphic implementation in terms of execution time.
The theoretical calculations of Loihi's energy consumption showed that it consumes $100 \text{ - } 1000$ times more energy than the FT accelerators.
Although the FT accelerators outperform the implementation of the \mbox{S-FT} architectures, we see the later as promising.

First, contrary to traditional DSPs, neuromorphic hardware design is still in its early stage,
and the newest \mbox{generations} of neuromorphic hardware are making big progress in terms of chip performance. In the case of Loihi, its newest version,
Loihi2\footnote{\url{https://download.intel.com/newsroom/2021/new-technologies/neuromorphic-computing-loihi-2-brief.pdf}},
promises spike generation and synaptic operations times that are $10$ and $5$ times faster, respectively.

Second, there are alternative neuromorphic designs that exploit the sub-threshold operation area and implement certain properties of the neuron models, such as the membrane potential, which uses analog voltages instead of digital values.
Some chips implementing these strategy outperform Loihi's $23.6\text{pJ}$ energy consumption per spike,
down to $0.381 \text{pJ}$ \cite{neckar2018braindrop}, $0.134 \text{pJ}$ \cite{moradi2017scalable}, or $0.077 \text{pJ}$ \cite{qiao2015reconfigurable}.
Adapting this hardware paradigm to the proposed SNN would potentially enable the design of neuromorphic solutions more efficient than FT accelerators.

Moreover, the spiking nature of the S-FT allows increasing the sparsity of the algorithm by discarding spikes that represent zero values.
The number of spikes that could be removed is highly dependent on the nature of the data and the error that can be sacrificed for improving the efficiency.
For the experiments that we have run, the spikes that occur at $t(x=0) \pm 5$ time steps account for $40-60\%$ of the total spikes.
Removing them would lead to a proportional reduction of the consumed energy and would not involve a significant loss in accuracy.
We would also like to emphasize that FT accelerators are specialized chips manufactured for a specific operation and with fixed parameters,
whereas neuromorphic counterparts are configurable chips used for research, so they are not optimized for this specific algorithm.


The existence of an SNN for performing the FT is also crucial for being able to implement sensor signal processing pipelines that are solely running in neuromorphic hardware.
This solution would avoid undesired costs introduced by the need to apply conversion stages between float numbers and spikes,
and install intermediate traditional DSPs for implementing this specific processing stage.
We anticipate that this algorithm will become more relevant as the SNN algorithms for processing higher level data get more mature.


\section{Conclusion}

In this work, we have proposed a neuron model for converting precise matrix multiplications into an SNN and tested the model for replicating the output of the FT.
Compared to previous works, we increased the sparsity of the network by proposing a novel neuron model that encodes the data using one spike per value.
In addition, we have reduced the total number of synaptic connections by designing an architecture based on the FFT algorithm.
We have also implemented the network in neuromorphic hardware and tested it with radar data obtained from real-world scenarios.

The experiments show that the error of the proposed algorithm is low in all scenarios.
However, the computational performance of the proposed work is behind FT accelerators in terms of energy consumption and execution time. 
We see this as a motivation for designing \mbox{ad hoc}  neuromorphic chips that are  highly specialized in performing the FT.
We hope that this work encourages further research on the design of novel hardware architectures that leverage the potential of neuromorphic hardware 
to create signal processing solutions faster and more efficient than current \mbox{state-of-the-art} solutions.
We believe that replacing the general-purpose board used for the experiments with a chip that implements a neuron model 
and a connection grid tailored for the proposed algorithm would significantly improve the computational
performance of the algorithm and would outperform the traditional solutions used today to implement the FT.

Besides the specific use case for computing the FT, the proposed SNN serves as a general model for tasks that involve matrix multiplications.
For instance, the proposed model could be applied for the conversion of deep convolutional neural nets into SNNs for inference after training.
This use case has already been explored for rate-based SNNs with minor losses in accuracy.
We believe that the proposed model would achieve a similar performance while improving the energy efficiency and execution time.
The main limitation is the elaborate parametrization of the threshold voltage, the amount of simulation steps, and the minimum and maximum boundaries of the encoding conversion.
Moreover, the implementation of the proposed model and other SNNs is restrained by the contemporary absence of commercial neuromorphic chips.

By means of this paper we aim to stimulate further research on the application of time-based SNNs for signal processing.
The S-FT can serve as the initial stage for larger processing pipelines, providing input for higher-level operations performed by other time-based SNNs,
such as object detection, tracking, or classification.


\section*{Acknowledgments}

This  research  has  been  funded  by  the  Federal  Ministry  of Education and Research of Germany in the framework of the KI-ASIC project
(16ES0995 and  16ES0996). 

The authors also acknowledge the financial support by the Federal Ministry of Education and Research of Germany in the programme of “Souverän. Digital. Vernetzt.”. Joint project 6G-life, project identification number: 16KISK001K.

We would like to thank Intel Corporation for granting us access to the chip Loihi and its associated resources.



\bibliographystyle{IEEEtran}
%

%





\clearpage


%

\appendix

\section{Theoretical framework of S-FT}
\thispagestyle{empty}







\subsection{Silent phase}
\label{sec:appendix_charging}

The DFT or FFT can be represented as matrix-vector multiplication.
Here, we present the mathematical proof that our neuron model can perform matrix-vector multiplications. 
After all spikes in the silent stage 1 occurred, we can state the final voltage of the neuron as

\begin{align}
    u_i = \sum_j w_{ij} (t_s - t_j).
\end{align}

Inserting the data-spike conversion results in 

\begin{align}
    u^{(1)}_i &= \sum_j w_{ij} (t_s - t_j) \\
        &= \sum_j w_{ij} (t_s - \frac{t_s}{2 x_\text{max}} \cdot (x_\text{max} - x_j)) \\
        &= \frac{t_s}{2 x_\text{max}} \sum_j w_{ij} x_j + \sum_j w_{ij} (t_s - \frac{t_s}{2 x_\text{max}} x_\text{max}) \\
        &= \frac{t_s}{2 x_\text{max}} \sum_j w_{ij} x_j + \sum_j w_{ij} (t_s - \frac{t_s}{2}) \\
        &= \gamma^{(1)} \sum_j w_{ij} x_j + \sum_j w_{ij} \frac{t_s}{2}.
\end{align}

By introducing the offset $b_i = - \sum_j w_{ij} \frac{t_s}{2}$ (that does not depend on the
input data) to the voltage, the voltage becomes

\begin{align}
    u^{(1)}_i &= \sum_j w_{ij} (t_s - t_j) + b_i^{(1)}\\
        &= \frac{t_s}{2 x_\text{max}}  \sum_j w_{ij} x_j\\
        &= \gamma^{(1)} \sum_j w_{ij} x_j \;.
\end{align}

Here, we can see that the voltage $u_i$ is proportional to the matrix-vector multiplication $u_i \sim \sum_j w_{ij} x_j$.

For calculating the voltage threshold we have to determine the maximum value that can be reached.
To maximize the scalar product $w_i \cdot x = \sum_j w_{ij} x_j$, we assume that $x$ has the same sign as $w_i$ and $|x_j| = x_\text{max} \forall j$.
Therefore, the maximum voltage of one neuron can be calculated as
\begin{align}
    u_{i, \text{max}} = \gamma^{(1)} \sum_j w_{ij} x_\text{max} \, .
\end{align}
For the threshold, we have to determine the maximum voltage of all neurons,
\begin{align}
   u^{(1)}_\text{max} = \max{u_{i, \text{max}}} = \max_{\forall i} \gamma^{(1)} x_\text{max} \sum_j w_{ij} \, .
\end{align}
\subsection{Spiking phase}

During the spiking phase a constant current $I_\text{ext} = \gamma^{(2)} = \frac{2 u_\text{max}}{t_s}$ is inserted. This value is chosen according to the data-spike conversion in (\ref{eq:ttfs}) and depends on the maximum voltage. By inserting this constant current into the voltage equation $u_i(t) = u_i + I_\text{ext} t$, the spike time can be calculated as

\begin{align}
    t_i &= \gamma^{(2)} (u_\text{max} - u_i) \\
        &= \gamma^{(2)}  (u_\text{max} - \gamma^{(1)} \sum_j w_{ij} x_j) \\
        &= \frac{t_s}{2} - \gamma^{(2)} \gamma^{(1)} \sum_j w_{ij} x_j \\
        &= \frac{t_s}{2} - \frac{t_s}{2 u_\text{max}} \frac{t_s}{2 x_\text{max}} \sum_j w_{ij} x_j \;.
\end{align}

It is important to highlight, that $u_\text{max}$ does not depend on the input data $x$ but only on the maximum value $x_\text{max}$,
the weights $w_{ij}$, and the time $t_s$. The voltage domain $u \in (-u_\text{max}, u_\text{max})$ is again symmetric.

\begin{align}
    u_\text{max} &= \frac{t_s}{2 x_\text{max}}  \max \sum_j | w_{ij} | \,,  \\
    u_\text{min} &= - \frac{t_s}{2 x_\text{max}}  \max \sum_j | w_{ij} | \\
\end{align}

Inserting the result in $t_i$ yields
 \begin{align}
    t_i &= \frac{t_s}{2} - \frac{t_s}{2 v_\text{max}} \frac{t_s}{2 x_\text{max}} \sum_j w_{ij} x_j \\
    &= \frac{t_s}{2} - \frac{t_s}{2 \frac{t_s}{2 x_\text{max}}  \max \sum_j | w_{ij} |} \frac{t_s}{2 x_\text{max}} \sum_j w_{ij} x_j \\
    &= \frac{t_s}{2} - \frac{t_s}{2 \frac{t_s}{2 x_\text{max}}  \max \sum_j | w_{ij} |} \frac{t_s}{2 x_\text{max}} \sum_j w_{ij} x_j \\
    &= \frac{t_s}{2} - \frac{t_s}{2 \max \sum_j | w_{ij} |}  \sum_j w_{ij} x_j \\
 \end{align}

For each layer, $l$ the output of the matrix-vector multiplication is normalized by $\max \sum_j | w^{(l)}_{ij} |$, which is independent of the input data.
These calculations can be repeated for each layer or matrix vector multiplication, respectively.

\section{Butterfly matrix}

$\textbf{W}_\text{8x8} \cdot \textbf{B}_\text{8x8} = $
\begin{align*}
&= \footnotesize\begin{pmatrix}
   \Re(\textbf{W}_\text{4x4}) & - \Im(\textbf{W}_\text{4x4})\\
   \Im(\textbf{W}_\text{4x4}) & \Re(\textbf{W}_\text{4x4})
   \end{pmatrix}
   \cdot 
   \begin{pmatrix}
   \Re(\textbf{B}_\text{4x4}) & - \Im(\textbf{B}_\text{4x4})\\
   \Im(\textbf{B}_\text{4x4}) & \Re(\textbf{B}_\text{4x4})
   \end{pmatrix}\\
   &= \scriptsize\left(\begin{array}{c c c c c c c c} 
   1 & 1 & 1 & 1 & 0 & 0 & 0 & 0 \\ 
   c_N^k & s_N^k & -c_N^k & -s_N^k & -s_N^k & c_N^k & s_N^k & -c_N^k \\ 
   c_N^{2 k} & -c_N^{2 k} & c_N^{2 k} & -c_N^{2 k} & -s_N^{2 k} & s_N^{2 k} & -s_N^{2 k} & s_N^{2 k} \\ 
   c_N^{3 k} & -s_N^{3 k} & -c_N^{3 k} & s_N^{3 k} & -s_N^{3 k} & -c_N^{3 k} & s_N^{3 k} & c_N^{3 k} \\ 
   0 & 0 & 0 & 0 & 1 & 1 & 1 & 1 \\ 
   s_N^k & -c_N^k & -s_N^k & c_N^k & c_N^k & s_N^k & -c_N^k & -s_N^k \\ 
   s_N^{2 k} & -s_N^{2 k} & s_N^{2 k} & -s_N^{2 k} & c_N^{2 k} & -c_N^{2 k} & c_N^{2 k} & -c_N^{2 k} \\ 
   s_N^{3 k} & c_N^{3 k} & -s_N^{3 k} & -c_N^{3 k} & c_N^{3 k} & -s_N^{3 k} & -c_N^{3 k} & s_N^{3 k} \\
   \end{array}\right)
\end{align*}


\end{document}